# The Use of Cuckoo Search in Estimating the Parameters of Software Reliability Growth Models

Dr. Najla Akram AL-Saati
Software Engineering Dept
College of Computer Sciences & Mathematics
Mosul, Iraq

Marwa Abd-AlKareem
Software Engineering Dept
College of Computer Sciences & Mathematics
Mosul, Iraq

*Abstract*— **this work aims to investigate the reliability of software products as an important attribute of computer programs; it helps to decide the degree of trustworthiness a program has in accomplishing its specific functions. This is done using the Software Reliability Growth Models (SRGMs) through the estimation of their parameters. The parameters are estimated in this work based on the available failure data and with the search techniques of Swarm Intelligence, namely, the Cuckoo Search (CS) due to its efficiency, effectiveness and robustness. A number of SRGMs is studied, and the results are compared to Particle Swarm Optimization (PSO), Ant Colony Optimization (ACO) and extended ACO. Results show that CS outperformed both PSO and ACO in finding better parameters tested using identical datasets. It was sometimes outperformed by the extended ACO. Also in this work, the percentages of training data to testing data are investigated to show their impact on the results.**

## I. INTRODUCTION

Software reliability has gained a huge importance recently due to the fact that it takes a long time for a company to build up a reputation for reliability, but only a short time to be acknowledged as "unreliable" subsequent to shipping a faulty product. Repeated appraisal of new product reliability and the constant reliability control of every shipped product are critical requirements in today's competitive business arena [1].

It has been found throughout the continual practice in this area that the process of proving or testing cannot assure absolute dependability neither on the product nor in its correctness. Therefore a metric is required to act as a measure for determining the degree of program correctness. One of the most widely used quality metrics is software reliability. It is usually measured using analytical models whose parameters are estimated from real failure data. [2]

Software Reliability can be defined as [3]:
"The probability that a system or product will perform in a satisfactory manner for a given period of time when used under specified operating conditions in a given environment."

To measure reliability in an effective manner, many issues have to be carefully considered, such as the accurateness of time to failure, the time to failures sequence, and failure mode data [4]. Evaluating software reliability requires many techniques, and nearly all of these techniques depend upon constructing prediction models with the ability to predict upcoming faults under diverse testing situations [5]. These models are generally named Software Reliability Growth Models (SRGMs).

In the past four decades, software reliability growth models began to be introduced for estimating the reliability of software programs. The nature of most of these models were non linear, which made the estimation of the parameters hard to accomplish using basic methods.[6] this idea opened the door for other methods to take part in estimating the parameters for non linear models, and the interest in Evolutionary Computation began to have effect in solving different Software Engineering problems.

Reliability growth models were deeply studies throughout the literature and their construction was investigated by many authors[2][7]. Numerous models were taken into consideration, and of the intensively used in the literature are those with two parameters, such as: Logarithmic [8], Exponential (Goel-Okumoto Model) [9], Power [10], S-Shaped (Yamada S-Shaped Model) [11][12] and Inverse Polynomial models [13].

Research in Software Engineering has recently been witnessing an enormous advance with the use of Evolutionary Computational methods. This is specially observed in finding acceptable solutions to prediction, estimation and optimization problems in Software Engineering. [14]

In this work, a number of models are introduced throughout the investigation. A detailed study of reliability and its growth models is introduced along with the problem of estimating their parameters, and then the methodology of the cuckoo search algorithm is presented to signify its role in estimating the parameters of the growth models. The results are compared with those obtained by [14] using the Exponential (Goel-Okumoto), the S-shaped and Power models whose parameters were estimated using PSO. In addition the results are also to be compared with those achieved by ACO and extended ACO in [6] using Exponential (Goel-Okumoto), S-shaped, Power, and M-O models. Comparisons indicate the efficiency of the CS algorithm.

## II. RELATED WORK

Parameter estimation in software reliability models were traditionally solved using the Maximum Likelihood method or the Least Square method; unfortunately these two methods are not suitable for non linear software reliability growth models. Therefore different solution methods and algorithms such as GAs, GP, NN, Fuzzy Logic, and Swarm Intelligence, have all had their share at trying to solve various problems in the Software Engineering area.

SRGMs were frequently studied and investigated throughout the literature; and here are some of these studies:

Kuo et al. in (2001) offered a framework for modeling software reliability by the use of various testing-efforts and fault detection rates. [7]. In (2002), Okamura et al. estimated a mixed Software Reliability Models using the Expectation-Maximization (EM) algorithm. [15] In addition, Okamura, Murayama and Dohi also used the (EM) algorithm in (2004) and developed a unified parameter estimation method for discrete software reliability models.[ 16]

By (2005), Huang [17], made a performance analysis of SRGMs with testing effort and change-point. After that, Ando, Okamura and Dohi introduced another work in (2006) about estimating Markov modulated software reliability models by the use of EM Algorithm [18]. In the same year, Sheta used PSO to solve the parameter estimation problem for the exponential, power and S-Shaped models [14]. During (2008), Bokhari, Quadri, and Khan, proposed the Exponetiated Weibull SRGM with various testing-efforts and optimal release policy with a performance analysis.[19] In (2009), Ohishi, Okamura, and Dohi presented a Non-Homogenous Poisson Process (NHPP) to develop an estimation algorithm of Gompertz software reliability model [20]. Throughout (2011), Quadri, Ahmad, and Farooq proposed a scheme for constructing a SRGM based on NHPP with generalized exponential testing – effort and optimal software release policy [21]. In the same year, Miglani and Rana proposed a greedy approach for ranking of different software reliability growth models [22].

Lately, in (2012), Shanmugam and Florence made a comparison of parameter best estimation methods and showed that ACO was the best among them [23]. At the same year, they made an improvement on ACO and compared it to their previous work [6].

Recently in (2013), a computational methodology based on weighted criteria was presented to the problem of performance analysis of various NHPP models [24]. In the same year, Okamura et.al, proposed a model based on a mixed gamma distribution, the estimation method was based on Bayesian estimation and the estimation algorithm was described by Markov chain Monte Carlo method with grouped data [25].

III. SOFTWARE RELIABILITY GROWTH MODELS (SRGMs)

*A. Definitions and Classification:*

Throughout the premature stages of developing and prototyping complex systems, reliability did not commonly meet customer requirements [1]. Reliability Models have usually been used to accurately evaluate and predict the behavior and performance of software reliability. In the 1970's, studies in software reliability models became more attracting and achieved greater advances, many reliability models have already been put into use.

Analysis methods of software reliability can either be white box or black box reliability analysis; these two differ from each other in that white box reliability analysis take into account the internal structure of the software to estimate its reliability, whereas black box (software reliability growth models) regards the software as a monolithic undividable unit by using failure data that took place in the middle of external interactions.[26] Fig. (1) shows the difference between white and black box reliability analysis and their relationship with software development stages.

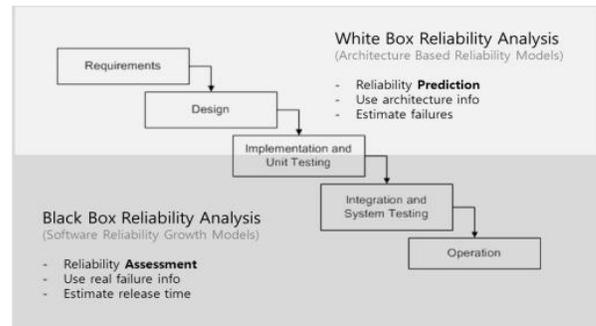

Figure 1. Black Box and White Box Reliability Analysis [26]

Software reliability models can generally be considered as static and the dynamic models. Static Models apply the modeling and analysis of program logic on the same code, while Dynamic models observe the temporary behavior of debugging process throughout the phase of testing [23].

There are many opinions related to the classification of reliability models. Basically there are two types [27]:

- **Defect Density Models** that try to predict reliability from design parameters. They employ code characteristics like code lines, loop nesting, external references, input/outputs, and others to estimate the number of defects in software.
- **Software Reliability Growth Models** which use test data to predict software reliability. These models intend to statistically correlate defect detection data with known functions like, for example, an exponential function. When that correlation is acceptable, the known function can be used to predict future behavior.

The use of reliability growth models (the focus of this work) for predicting software reliability signifies a huge challenge for software testing. predicting the number of faults inhabited in software programs gives a considerable assists in denoting the day for software release and control project resources (people and money) [28]. Most software reliability growth models provide work for the estimation of two or three parameters, these models include the predictable number of failures in the software, and the initial failure intensity.

*B. Characteristics and Notations:*

The characteristics of Software Reliability Growth Model's should be satisfied for the Software Reliability Model [29]. These Characteristics are as follows:

- SRGM can be viewed as a product of a cumulative density function and a positive constant.

$$H(t) = a\left[1 - \exp\left(-\int_0^1 d(x)\,dx\right)\right] = aG(t), \quad (1)$$

Where:
H (t) is mean value function.
G (t) is Cumulative Density Function.

- The fault detection rate must be finite, and G(t) must meet the condition that the corresponding failure rate function is finite.

$$h(t) = d(t)[a - H(t)] = ag(t), \qquad (2)$$

Where
g(t) = dG(t)/dt, the probability density function associated with G (t).

$$d(t) = ag(t)\frac{ag(t)}{a - aG(t)} = \frac{g(t)}{1 - G(t)} = r(t) \qquad (3)$$

Where:
r(t) the failure rate function associated with G (t).
h(t): is intensity function.

- It is necessary to appropriately represent the right tail of g(t) behavior to achieve a good reliability prediction. The right tail of SRGM associated with g(t) should be heavy.

The terms in Table (I) are to be used in defining the Models in the next subsections.

TABLE I. TERMS USED FOR DEFINING SRGMs

| Term | Definition |
|---|---|
| μ(t) | Denotes the mean failure function, i.e., the expected number of failures observed over a period of time t. |
| λ(t) | Denotes the failure intensity function, i.e., failure rate |
| a | The initial estimate of the total failure recovered at the end of the testing process. |
| b | Represents the ratio between the initial failure intensity λ0 and total failure. |
| NHPP | The Non Homogenous Poisson Process: provides probability that the number of failures at a time t will have a particular value. |

*C. Models Employed in this Work:*

In this work a number of models are considered, in particular those that are frequently and commonly referenced in the literature.

The models studied in this work are:

*1) Exponential Model (Goel-Okumoto G-O) [9]:* Goel-Okumoto model is recognized as a finite failure model that can be modeled as NHPP. The Prediction of the model can be given as:

$$\mu(t) = a(1 - e^{-bt}) \qquad (4)$$

$$\lambda(t) = abe^{-bt} \qquad (5)$$

The number of faults to be detected (a) is handled as a random variable for which the observed value depend on the test and other environmental factors. This is fundamentally different from other models that treat the number of faults as a fixed unknown constant [30].

*2) The Power Model [10]:*
This model has the objective of computing the reliability of hardware systems during testing process. It is also based on the NHPP [31]. The equations (6) and (7) rule the relationship between the time t and both $\mu$ (t) and $\lambda$ (t).

$$\mu(t) = at^b \qquad (6)$$

$$\lambda(t) = abte^{b-1} \qquad (7)$$

*3) The Yamada Delayed S-Shaped Model [11][12]:*
The Delayed S-Shaped Model is of the gamma distribution class. But the number of failures per time period is a Poisson type with the use of the classification scheme of Musa and Okumoto rather than considered as Binomial. This model describes the software reliability process as a delayed S-shaped model. The model represents a learning process since some improvement was added to the exponential model based the growing experience of the project team. This model is also a finite failure model [14]. The system equation for $\mu$ (t) and $\lambda$ (t) are:

$$\mu(t) = a(1 - (1 + bt)e^{-bt}) \qquad (8)$$

$$\lambda(t) = ab^2 t^{-bt} \qquad (9)$$

*4) Musa-Okumoto Logarithmic Model [8]:*
This is continuous time-independently distributed inter failure time model. This is a modification of the J-M model where a geometrically decreasing hazard function is introduced, considering that only one fatal error is removed during each debugging interval. Faults are not removed until the occurrence of a fatal one at which time the accumulated group of faults is removed. The hazard function after a restart is a fraction of the rate which was attained when the system crashed [23]. The Prediction Model form is given as:

$$\mu(t) = a \ln(1 + bt) \qquad (10)$$

*D. Parameter Estimation:*

Estimating the parameters problem for nonlinear systems can be stated and formulated as a function optimization problem. The purpose is to discover a set of parameters that provide the best fit to a measured data based on a specific type of function to be optimized. Such parameters are found using a search procedure in the space of values specified in advance. Searching techniques are bound to the complexity of the search space, and the use of Gradient search might find local minimum solution but not optimal ones. Stochastic search algorithms, on the other hand, such as Evolutionary Algorithms present a more reliable functionality in estimating models' parameters. [14]

Swarm Intelligence has been successfully used to provide efficient mechanisms in searching for solutions to this problem, of these mechanisms, PSO and ACO were used as stated previously and in this work Cuckoo Search (CS) is applied to search for better estimates associated with the parameter values of SRGMs and its efficiency and performance is compared to those obtained by PSO and ACO.

## IV. CUCKOO SEARCH (CS):

New meta-heuristic search algorithms are rapidly increasing under the paradigm of swarm intelligence, resembling the intelligence inhabited in creatures from nature. Algorithms such as Particle Swarm Optimization (PSO), Ant colony optimization (ACO), Bee Colony Optimization (BCO) Fish Schools, and many others have become well known in solving various problems. Of these algorithms, Cuckoo Search (CS) [32] which was developed by Yang and Deb in 2009, has proven to be very promising through preliminary studies and conformed giving more robust and precise results than PSO and ABC (Artificial Bee Colony).[33]

### A. Cuckoo's Inspiring Behavior:

Some species of Cuckoo are interesting, such as the Ani and Guira, as they have a strange habit of laying their eggs in public nests; they may also remove others' eggs to boost the hatching probability of their own eggs [34]. Quite a number of species engage the obligate brood parasitism by laying their eggs in the nests of other host birds (often other species). There are basically three types of brood parasitism:

- Intra-specific brood parasitism.
- Cooperative breeding.
- Nest takeover.

Some host birds can go on a direct conflict with the intruding cuckoos. If a host bird finds that the eggs are not its own, it will either throw away these alien eggs or simply dump its nest and build a new one in another place. Some cuckoo species such as the new world brood-parasitic Tapera have evolved somehow that female parasitic cuckoos are often very expert in the mimicry in the egg's color and pattern of a few chosen host species. This decreases their eggs' probability of being abandoned and therefore enhances their reproduction activity. [32]

Also taking into account that parasitic cuckoos habitually choose a nest where the host bird just laid its own eggs. Usually, the cuckoo eggs hatch a little earlier than their host eggs. When the first cuckoo baby bird is hatched, its first instinct act will be to throw out host eggs, this will increase its share of food that is supplied by the nest's host bird. Various Studies in cuckoo behavior also indicate that a cuckoo baby bird can mimic the call of host baby bird to obtain more feeding opportunity.[32]

In this paper, CS will be applied to parameter estimation of SRGMs to conduct a more detailed study of its characteristics and to verify it against benchmark datasets. After that a discussion is establish the unique features of Cuckoo Search and propose topics for further studies.

### B. Lévy Flights:

Animals habitually search for food in a random or quasi-random style. Generally, the foraging path of an animal is in fact a random walk; this is because the next move depends on the current location or state and the transition probability to the next location. The chosen direction depends implicitly on a probability which can be mathematically modeled. For instance, various studies on animals and insects have shown that their flight behavior demonstrates the typical characteristics of Lévy flights [35].

Reynolds and Frye [35] explored fruit flies (Drosophila Melanogaster), and indicated that they explore their landscape using a series of straight flight paths punctuated by a sudden 90o turn, leading to a Lévy-flight-style intermittent scale-free search pattern. Afterwards, Studies on human behavior were conducted as well and also showed the typical feature of Lévy flights. Many other things can also be related to Lévy flights, such as light [36]. That is why such behavior has been applied to optimization and optimal search, where preliminary results show its promising capability [37].

### C. Search Strategy for Cuckoos:

For the ease of description, the following three idealized rules are used [37]:

- In each time, every cuckoo lays one egg in a randomly chosen nest.
- Only best nests having high quality eggs (solutions) will continue to the next generations;
- The available host nests are fixed in number. A host can discover an alien egg with a probability pa ∈ [0, 1]. When discovered, the host bird can either throw the egg away or dump the nest to build a totally new one in another location.

```
Cuckoo Search via Lévy Flights
Begin
    Objective function f(x), x = (x1, ..., xd) T
    Generate initial population of n host nests xi (i = 1, 2, ..., n)
    While (t <MaxGeneration) or (stop criterion)
    Get a cuckoo randomly by Lévy flights
        Evaluate its quality/fitness Fi
    Choose a nest among n (say, j) randomly
    If (Fi > Fj),
            Replace j by the new solution;
    End
    A fraction (pa) of worse nests is abandoned and new ones are built;
    Keep the best solutions (or nests with quality solutions);
    Rank the solutions and find the current best
    End while
    Post-process results and visualization
End
```

Figure 2. Pseudo code of the Cuckoo Search (CS) [32]

Based on these three rules, the basic steps of the Cuckoo Search (CS) can be presented as the pseudo code shown in Fig. 2 [32]. To give more simplicity, the last assumption can be approximated by a fraction pa of the n nests being replaced by new nests (with new random solutions at new locations). The fitness can be defined in the same way as done in Genetic Algorithms. When a new solution $x_i^{(t+1)}$ is generated for the i'th cuckoo, a Lévy flight is done as the following:

$$x_i^{(t+1)} = x_i^{(t)} + \alpha \oplus \text{Lévy}(\lambda), \qquad (11)$$

where:

$\alpha > 0$ is the step size (it should be related to the scales of the problem at hand). Most of the time, it is used as $\alpha = O(1)$.

$\oplus$ is the entry-wise multiplications.

Lévy flight is used to conduct a random walk drawn from a Lévy distribution for large steps:

$$\text{Lévy} \sim u = t^{-\lambda}, (1 < \lambda \leq 3), \quad (12)$$

This has an infinite variance with an infinite mean. The successive jumps/steps of a cuckoo basically form a random walk which obeys a power-law step-length distribution with a heavy tail.

## V. PARAMETER ESTIMATION BASED ON CS:

Based on the characteristics of the SRGMs, the cuckoo search is implemented as follows:

- To solve the problem of parameter estimation, the principal of Root Mean Square Error (RMSE) is used as in (13):

$$\text{RMSE} = \sqrt{\frac{1}{N}\sum_{t=1}^{N}[m(t) - \mu(t)]^2}, \quad (13)$$

The total actual discovered failure number in time t, is m(t), the predicted failure number by SRGM are expected failure number by time t is μ(t).

- each nest carries a pair of eggs representing solution parameters (a, b), at each time a cuckoo is chosen according to levy flight, its fitness is evaluated and compared to the fitness of a randomly chosen nest, the best fitness is kept.
- A fraction (pa) of nests is abandoned and new nests are built.

## VI. TESTS AND RESULTS

To test the efficiency of the search algorithm employed in this work, two types of testing are conducted. In the beginning, comparisons are made with previous results gained using PSO (with three models) and ACO (with four models) using the same related datasets. Then, the splitting of Datasets between training and test sets is investigated to show the impact of their sizes on results; this is done using three models and one benchmark dataset. Table (II) shows the parameter settings for the cuckoo search employed in this work for all the experiments.

TABLE II. PARAMETER SETTINGS FOR THE CUCKOO SEARCH ALGORITHM

| Parameter | Value |
|---|---|
| Lower and Upper bounds for Parameter a | [0.00001 - 2000] |
| Lower and Upper bounds for Parameter b | [0.00001 – 1] |
| Number of Cuckoos | 1 |
| Number of Nests | 10 |
| Number of Eggs | 2 |
| Number of iterations (Generations) | 100 |
| Alpha | 0.01 |
| Discovery rate | 0.25 |

### A. Experimental Data used in this work

Datasets used in this work are chosen in accordance to those referenced by other researchers with which the comparisons were made; the first group of datasets (compared with PSO) is taken from [14] for Data1, Data2, and Data3. The second group (compared with ACO and extended ACO) is selected from The Software Reliability Dataset which was compiled by John Musa of Bell Telephone Laboratories [38] for Project 2, Project 3, and Project 4.

### B. Comparison with other Swarm Algorithms:

The cuckoo search algorithm's methodology is compared to that of PSO using the same datasets employed in his work, and Table (III) signifies the training and testing percentages as divided by Sheta [14] for exponential (EXP), power (POW), and Delayed S-Shaped Yamada Model (DSS), the results show the clear improvement achieved in testing the parameters for the specified models for Data1. Tables (IV) and (V) shows the comparisons made using Data2 and Data3 for the same models.

These results were accomplished using 100 iterations when compared to that used by [14], where 1000 iteration were required to achieve the given PSO results.

TABLE III. COMPARISON WITH PSO USING DATASET (DATA1)

| Search Model | RMSE - Training 70% | | RMSE - Testing 30% | |
|---|---|---|---|---|
| | PSO | CS | PSO | CS |
| EXP(G-O) | 20.2565 | 34.0933 | 119.4374 | 16.8945 |
| POW | 22.2166 | 44.8663 | 152.9372 | 33.6623 |
| DSS | 15.9237 | 32.6376 | 26.3015 | 10.9945 |

TABLE IV. COMPARISON WITH PSO USING DATASET (DATA2)

| Search Model | RMSE - Training 70% | | RMSE - Testing 30% | |
|---|---|---|---|---|
| | PSO | CS | PSO | CS |
| EXP(G-O) | 24.9899 | 33.2311 | 80.8963 | 14.2998 |
| POW | 32.3550 | 47.0571 | 149.9684 | 56.6807 |
| DSS | 20.8325 | 27.9159 | 17.0638 | 11.8833 |

TABLE V. COMPARISON WITH PSO USING DATASET (DATA3)

| Search Model | RMSE - Training 70% | | RMSE - Testing 30% | |
|---|---|---|---|---|
| | PSO | CS | PSO | CS |
| EXP(G-O) | 12.8925 | 13.5404 | 13.6094 | 8.9523 |
| POW | 11.9446 | 13.0886 | 14.0524 | 13.4669 |
| DSS | 18.5807 | 13.6634 | 47.4036 | 15.1916 |

The methodology is further compared to ACO [23] and extended ACO [6] using the same datasets used in their work, they used the whole sets of data for training. Tables (VI), (VII), and (VIII) indicate the training percentages for Goel-Okumoto (G-O), power (POW), Delayed S-Shaped Yamada Model (DSS), and the Musa-Okumoto (M-O) model for Projects 2, 3, and 4.

Results in Table (VI) show that using Project2 with (G-O, POW, and DSS) models, CS outperformed ACO but not EX-ACO. But for the (M-O) model, CS gave the worst results.

TABLE VI. COMPARISON WITH ACO AND EXTENDED ACO (PROJECT2)

| Search Model | Project2 – Training 100% | | |
|---|---|---|---|
| | ACO | Ex-ACO | CS |
| G-O | 60.0371 | 28.5891 | 41.7971 |
| POW | 52.8854 | 34.0521 | 45.9783 |
| DSS | 52.8854 | 33.0461 | 42.2256 |
| M-O | 26.0385 | 17.359 | 41.7732 |

For Project3, CS outperformed both ACO and EX-ACO for all models as illustrated in Table (VII). Using Project4, Table (VIII) indicates that CS surpassed both ACO and EX-ACO for (G-O, POW, and DSS) models. As for the M-O model, CS performed better than ACO but not better than EX-ACO. Fig. 3 to 5 depicts the data in Tables (VI, VII, and VIII) to indicate the differences among the three datasets used for the same models.

TABLE VII. COMPARISON WITH ACO AND EXTENDED ACO (PROJECT3)

| Search Model | Project3 – Training 100% | | |
|---|---|---|---|
| | *ACO* | *Ex-ACO* | *CS* |
| *G-O* | 71.5489 | 34.0709 | **21.7256** |
| *POW* | 57.5801 | 47.5814 | **15.5885** |
| *DSS* | 57.5801 | 48.4914 | **22.4944** |
| *M-O* | 36.1891 | 24.126 | **19.5448** |

TABLE VIII. COMPARISON WITH ACO AND EXTENDED ACO (PROJECT4)

| Search Model | Project4 – Training 100% | | |
|---|---|---|---|
| | *ACO* | *Ex-ACO* | *CS* |
| *G-O* | 71.4015 | 35.0007 | **25.7682** |
| *POW* | 53.2234 | 34.2645 | **28.1951** |
| *DSS* | 53.2234 | 35.2635 | **25.7294** |
| *M-O* | 33.1728 | **22.1152** | 26.4575 |

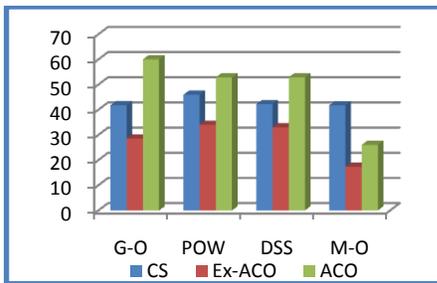

Figure 3. Diference among the Three Search Algorithms (Project2)

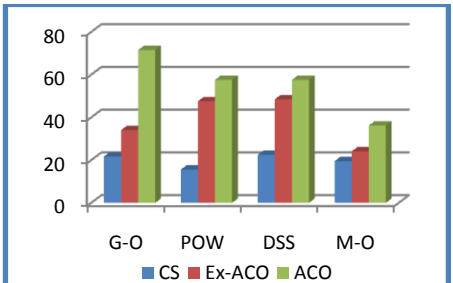

Figure 4. Diference among the Three Search Algorithms (Project3)

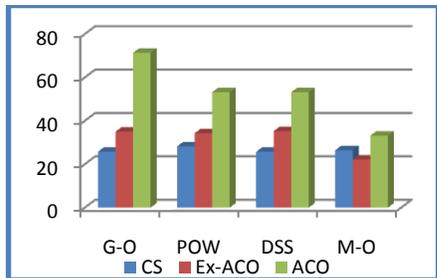

Figure 5. Diference among the Three Search Algorithms (Project4)

As a result, CS can perform better than PSO in all cases considered in this study. For ACO, it was noticed that CS has achieved better results in most cases, and for some cases it also surpassed EX-ACO. This is largely due to the nature of the datasets used in correlation with the model employed in the testing of parameters.

*C. The Impact of Training and Testing Data:*

Through the process of parameter estimation, the datasets used can either be used just for training, or it can be divided into two sets: Training and Testing. This division can have a large influence on results, and that is way it is investigated in this subsection.

The study included the three datasets (Data1, Data2, and Data3) used in the previous subsection. Tables (IX), (X), and (XI) show the impact of each model using the three datasets and five different percentages for training and testing for models (EXP, POW, and DSS). From the results, it can be seen that increasing the training data percentage over testing data, will enhance testing results and worsen training results, and vise versa. This is quite logical, as feeding a large amount of data to the training process forces the search procedure (cuckoo search here) to find very few but excellent solutions (proved by the corresponding testing process). When smaller amounts of data are used for training, they become unsatisfactory, and a large amount of testing data is not required.

To overcome this problem of overestimation and underestimation, a balanced point is chosen such as (70%, 30%) to give training reasonable amounts of data to achieve good training and adequate data to the testing process.

TABLE IX. IMPACT OF TRAINING AND TESTING PERCENTAGES (DATA1)

| **Model** | **EXP** | **POW** | **DSS** |
|---|---|---|---|
| *Training 90%* | 38.062 | 61.940 | 41.833 |
| *Testing 10%* | 8.811 | 6.481 | 2.747 |
| *Training 80%* | 36.37 | 54.77 | 35.77 |
| *Testing 20%* | 10.703 | 16.335 | 7.113 |
| *Training 70%* | **34.093** | **44.866** | **32.638** |
| *Testing 30%* | **16.895** | **33.662** | **10.995** |
| *Training 60%* | 32.877 | 39.858 | 30.985 |
| *Testing 40%* | 22.253 | 48.956 | 21.555 |
| *Training 50%* | 27.443 | 27.877 | 24.760 |
| *Testing 50%* | 31.539 | 73.668 | 36.336 |

TABLE X. IMPACT OF TRAINING AND TESTING PERCENTAGE (DATA2)

| **Model** | **EXP** | **POW** | **DSS** |
|---|---|---|---|
| *Training 90%* | 40.374 | 62.153 | 30.843 |
| *Testing 10%* | 3.490 | 7.574 | 6.209 |
| *Training 80%* | 36.841 | 56.576 | 28.637 |
| *Testing 20%* | 7.126 | 25.381 | 10.598 |
| *Training 70%* | **33.231** | **47.057** | **27.916** |
| *Testing 30%* | **14.299** | **56.680** | **11.883** |
| *Training 60%* | 29.876 | 39.468 | 25.648 |
| *Testing 40%* | 33.402 | 86.050 | 45.975 |
| *Training 50%* | 26.377 | 33.046 | 24.907 |
| *Testing 50%* | 68.858 | 125.906 | 71.220 |

TABLE XI.  IMPACT OF TRAINING AND TESTING PERCENTAGE (DATA3)

| Model | EXP | POW | DSS |
|---|---|---|---|
| *Training 90%* | 18.539 | 19.525 | 17.229 |
| *Testing 10%* | 1.613 | 3.256 | 5.158 |
| *Training 80%* | 15.299 | 17.311 | 15.489 |
| *Testing 20%* | 3.916 | 8.287 | 10.588 |
| *Training 70%* | **13.540** | **13.089** | **13.663** |
| *Testing 30%* | **8.952** | **13.467** | **15.191** |
| *Training 60%* | 12.484 | 12.986 | 11.874 |
| *Testing 40%* | 16.476 | 16.578 | 33.636 |
| *Training 50%* | 11.823 | 12.132 | 10.673 |
| *Testing 50%* | 34.240 | 20.012 | 60.926 |

Also in this work, the number of iterations (generations), the number of nests (population size) and the probability (pa) were varied to find the best suitable settings. Number of iterations was varied from (50 to 1000) and the value 100 iteration was quite capable of achieving the best results. Nests of (10, 15, 20, 25, 50, 100, 150, and 500) were tried and the sufficient value was found to be (10). As for the probability pa, values of (0, 0.01, 0.02, 0.05, 0.1, 0.15, 0.2, 0.25, 0.5, and 0.7) were used and the best setting was found to be (0.25).

## VII. CONCLUSIONS AND FURTHER RECOMMENDATIONS

It has been found through this work that the CS algorithm can be successfully applied to find good and acceptable solutions to the problem of parameter estimation of Software Reliability Growth Models. Models considered in this work are: the Exponential, Power, S-Shaped, and M-O models. The search strategy of the cuckoo can efficiently navigate throughout the search space of the problem and locate very good solutions using fewer iterations and smaller populations.

First, the Results were compared to PSO, ACO, and Extended ACO; the results clearly outperformed PSO, they were better than ACO in almost all cases, but were sometime worse than the extended ACO.

Second, the testing and training data sizes were investigated, pairs of (training, testing) were considered using three datasets they were: (90%, 10%), (80%, 20%), (70%, 30%), (60%, 40%), and (50%, 50%). It is concluded that increasing the percentage of training data, makes training hard and testing very simple and not sufficient enough. On the contrary, when percentage of training is decreased, the training data becomes insufficient to train and the testing data becomes very large and unnecessary. Thus a counterbalance point is chosen in the middle (70%, 30%) to give training a justified amount of data to accomplish good training and fair enough data to the testing process.

As for further recommendations, other swarm intelligent methods can be applied and compared; this may uncover more suitable parameter estimation methods. Future work might also include the application of other evolutionary search methods to construct new SRGMs that can assess the reliability more adequately.